\documentclass{article}
\usepackage{amssymb}
\usepackage{amsmath}
\usepackage{algorithm}
\usepackage{algpseudocode}
\usepackage{enumitem}
\usepackage{pifont}
\usepackage[final]{corl_2025}
\usepackage{hyperref}
\usepackage{times}
\usepackage{bbm}
\usepackage{multicol,epigraph,wrapfig}
\usepackage{graphics,graphicx,caption,float,subcaption,booktabs,multirow,array,ifthen,tabu,dblfloatfix,url,xparse,mathtools,patchcmd,xspace,nicefrac,microtype,amsmath,amsfonts,bm,ragged2e,stackengine,etoolbox,xpatch,enumerate,xstring,setspace,tabularx,makecell,changepage,cuted,titlesec,wrapfig, sidecap}
\usepackage[para]{footmisc}

\title{Deep Reactive Policy: Learning Reactive Manipulator Motion Planning for Dynamic Environments}
\author{\textbf{Jiahui Yang$^{\ast}$\qquad Jason Jingzhou Liu$^{\ast}$\qquad Yulong Li} \\
\textbf{Youssef Khaky\qquad Kenneth Shaw\qquad Deepak Pathak}\\
Carnegie Mellon University\\ 
\vspace{-4mm}
$^{\ast}$Equal contribution}

\begin{document}
{\setstretch{1.25}\maketitle}


\begin{figure}[h]
    \includegraphics[width=1\linewidth]{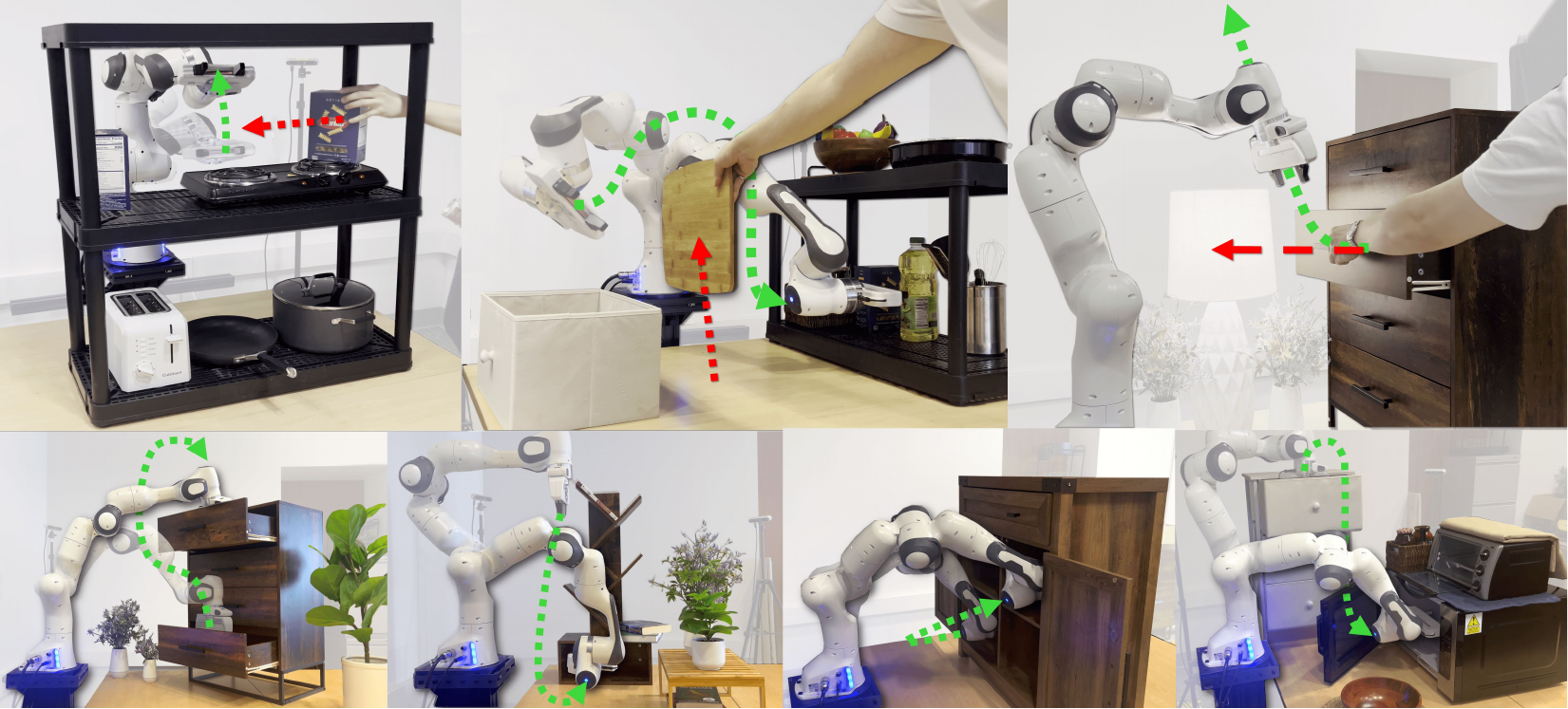}
    \centering
    \captionof{figure}{\small We present Deep Reactive Policy (DRP), a point cloud conditioned motion policy capable of performing reactive, collision-free goal reaching in diverse complex and dynamic environments.}
    \label{fig:teaser}
    \vspace{-0.1in}
\end{figure}

\vspace{7mm}

\begin{abstract}
Generating collision-free motion in dynamic, partially observable environments is a fundamental challenge for robotic manipulators. Classical motion planners can compute globally optimal trajectories but require full environment knowledge and are typically too slow for dynamic scenes. Neural motion policies offer a promising alternative by operating in closed-loop directly on raw sensory inputs but often struggle to generalize in complex or dynamic settings.

We propose Deep Reactive Policy (DRP), a visuo-motor neural motion policy designed for reactive motion generation in diverse dynamic environments, operating directly on point cloud sensory input. At its core is IMPACT, a transformer-based neural motion policy pretrained on 10 million generated expert trajectories across diverse simulation scenarios. We further improve IMPACT's static obstacle avoidance through iterative student-teacher finetuning. We additionally enhance the policy's dynamic obstacle avoidance at inference time using DCP-RMP, a locally reactive goal-proposal module.

We evaluate DRP on challenging tasks featuring cluttered scenes, dynamic moving obstacles, and goal obstructions. DRP achieves strong generalization, outperforming prior classical and neural methods in success rate across both simulated and real-world settings. Video results and code available at \href{https://deep-reactive-policy.com/}{deep-reactive-policy.com}
\end{abstract}


\newpage
\section{Introduction}

To operate in natural human environments like homes and kitchens, robots must navigate safely in fast-changing, partially observable settings. This demands an intuitive understanding of robot's own physical presence within the world. At the core of this capability is collision-free motion generation. Motion generation can operate beneath high-level policy layers such as VLMs or behavior cloning agents, ensuring that the robot's actions remain both safe and physically feasible.

To achieve this, robots have traditionally relied on motion planning approaches.  Search-based methods, such as A* \cite{a*} and AIT* \cite{ait*}, are capable of finding globally optimum solutions, but they assume complete knowledge of the environment and static conditions. Due to their long execution times, these planners are typically run offline to generate fixed open-loop trajectories that the robot executes, limiting their performance in avoiding dynamic obstacles. Reactive controller-based approaches~\cite{khatib1985real, rmp, fabrics, bhardwaj2022storm}, such as Riemannian Motion Policies (RMP) \cite{rmp} and Geometric Fabrics ~\cite{fabrics}, offer reactive collision avoidance in dynamic scenes.  However, these approaches lack global scene awareness and often become trapped in local minima in complex environments \cite{mpinet}.

An alternative is to formulate motion generation as a visuo-motor neural policy that maps raw visual observations directly to actions. Unlike open-loop planners, a learned visuo-motor policy continuously processes live sensory inputs—such as point clouds—to adapt its behavior on the fly without requiring known models of the scene. This real-time closed loop adaptability is crucial in scenarios where the goal becomes temporarily obstructed by dynamic obstacles or during sudden environmental changes.

Generating such visuo-motor neural policies requires a robust training methodology. Several works have proposed using traditional motion planners to produce ground truth trajectories to serve as supervision for training~\cite{mpnet, mpinet, nmp, mpiformer}. Unfortunately, prior methods such as M$\pi$Net~\cite{mpinet} have only demonstrated limited success in simple settings and often fail to generalize to unseen environments, constraining their broader applicability. More recent efforts, like NeuralMP~\cite{nmp}, attempt to address these generalization issues by introducing test-time optimization to correct for policy inaccuracies. While this improves accuracy, it requires runtime search before execution, sacrificing the reactivity necessary for fast-changing environments.

Our method, Deep Reactive Policy (DRP), is a visuo-motor neural motion policy designed for reactive motion generation in diverse dynamic environments, operating directly from point cloud sensory input. At the core of DRP is IMPACT (Imitating Motion Planning with Action-Chunking Transformer), a transformer-based neural motion policy.  First, we pretrain IMPACT on 10 million motion trajectories generated in diverse simulation environments leveraging cuRobo \cite{curobo}, a state-of-the-art GPU-accelerated motion planner. Next, we finetune IMPACT with an iterative student-teacher method to improve the policy's static obstacle avoidance capabilities. Finally, we integrate a locally-reactive goal-proposal module, DCP-RMP, with IMPACT to further enhance dynamic reactivity.  Altogether, we call our approach Deep Reactive Policy (DRP), enabling the policy to generalize from learned experiences and develop an intuitive understanding of the changing environment.

Our core contributions are:
\begin{itemize}[leftmargin=1.5em]
    \item We scale up motion data generation to train IMPACT, a novel end-to-end transformer-based neural motion policy conditioned on point cloud observations.
    \item We further improve IMPACT's obstacle avoidance performance via finetuning with iterative student-teacher distillation.
    \item We enhance IMPACT's dynamic obstacle avoidance performance via a locally reactive goal-proposal module, DCP-RMP.
\end{itemize}
We evaluate DRP on both simulation and real-world environments, featuring complex obstacle arrangements and dynamic obstacles. DRP consistently outperforms previous state-of-the-art motion planning methods, as detailed in Section \ref{sec:exp}.
\section{Related Work}
\label{sec:citations}
\vspace{-0.05in}
\subsection{Global Planning Methods}
\vspace{-0.05in}
\label{sec:global_planning}
Global planners generate collision-free trajectories by exploring the full state space, offering asymptotic completeness and optimality. Search-based methods like A* ~\cite{a*} and its variants~\cite{likhachev2003ara,likhachev2005anytime,koenig2006new} guarantee optimality under admissible heuristics but scale poorly in high-dimensional continuous domains due to discretization. Sampling-based planners such as PRM~\cite{prm}, RRT~\cite{rrt}, and their extensions~\cite{lazyprm,lavalle2001rapidly,kuffner2000rrt,bit*,ait*} improve scalability and efficiency through continuous-space sampling and informed exploration. Trajectory optimization methods~\cite{zucker2013chomp,schulman2014motion,dragan2011manipulation} refine trajectories via continuous optimization but are sensitive to initialization and local minima. Recent work, cuRobo, advances trajectory optimization with GPU parallelization. However, it still plans from scratch for each new problem and struggles with noise and partial observability, limiting its applicability in real-world and dynamic environments. In contrast, our method overcomes these challenges by learning from diverse planning data and directly generating actions from raw point cloud inputs.

\subsection{Locally-Reactive Methods}
\label{sec:local_reactive}

Locally reactive controllers generate collision-free motion by steering toward goals while avoiding nearby obstacles~\cite{khatib1985real, rmp, fabrics, bhardwaj2022storm}. Though effective in dynamic settings, they often get trapped in local minima within cluttered environments~\cite{mpinet}. We address this by training a neural policy on globally optimal planners, enabling it to produce globally aware motion even in complex scenes while retaining closed-loop reactivity.

\subsection{Learning-based Methods}
\label{sec:learning_based}
Neural networks run efficiently at inference time, and have been widely used to accelerate motion planning. A line of work augments classical motion planners with learned components that bias sampling~\cite{mpnet,kumar2019lego,zhanglearning,chamzas2021learning}, guide search~\cite{belt,compnet} or initialize optimization~\cite{diffusionseeder}. These hybrid approaches preserve the robustness of classical planning while leveraging learning to improve sample efficiency and planning speed, particularly in high-dimensional or constrained scenarios. In contrast, more recent methods~\cite{dynmpnet, mpdiffusion, edmp} generate feasible motion directly, handling dynamic constraints, capturing multimodal distributions, and incorporating scene-specific costs without explicit planning at inference. 

While effective, these methods often rely on ground-truth states, known obstacle geometry, or open-loop trajectory prediction without closed-loop reactivity. To overcome these limitations, recent approaches develop end-to-end policies that operate directly on point clouds, facilitating real-world deployment without the need to perform scene reconstruction. M$\pi$Nets~\cite{mpinet} and M$\pi$Former~\cite{mpiformer} use supervised learning on point cloud inputs, achieving strong in-distribution results, but struggling to generalize due to limited training diversity. NeuralMP~\cite{nmp} improves generalization through scene and obstacle randomization, but relies on slow test-time optimization, limiting real-time performance. Our method addresses both issues, providing robust generalization and fast closed-loop execution.
\vspace{-0.05in}
\section{Method}
\label{sec:method}
\begin{figure}[t]
    \centering
    \includegraphics[width=0.99\textwidth]{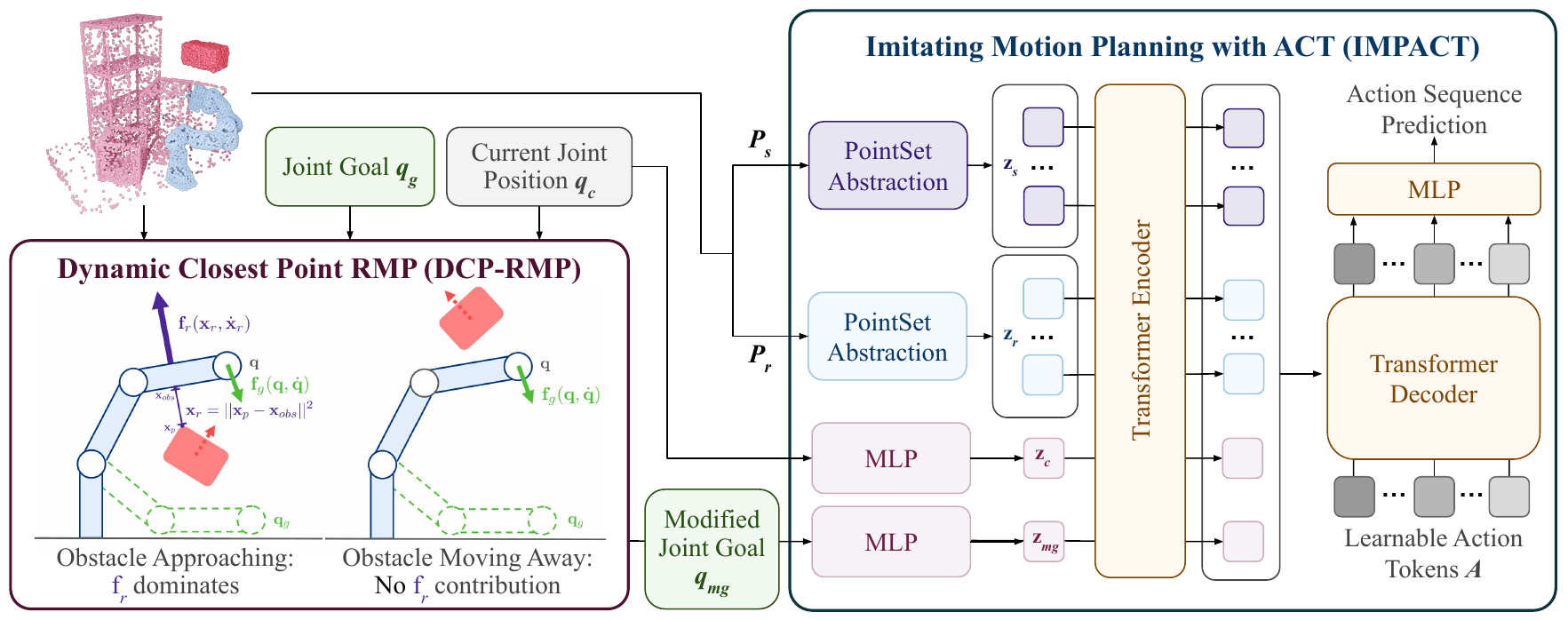}
    \caption{Deep Reactive Policy (DRP) is a visuo-motor neural motion policy designed for dynamic, real-world environments. First, the locally reactive DCP-RMP module adjusts joint goals to handle fast-moving dynamic obstacles in the local scene. Then, IMPACT, a transformer-based closed-loop motion planning policy, takes as input the scene point cloud, the modified joint goal, and the current robot joint position to output action sequences for real-time execution on the robot.}
    \label{fig:method}
    \vspace{-1.2em}
\end{figure}

\vspace{-0.05in}
We present Deep Reactive Policy (DRP), a neural visuo-motor policy that enables collision-free goal reaching in diverse, dynamic real-world environments. An overview of the full system architecture is shown in Figure~\ref{fig:method}.

At the core of DRP is IMPACT, a transformer-based policy that generates joint position targets conditioned on a joint-space goal and live point-cloud input. IMPACT is trained in two phases. First, it undergoes pretraining via behavior cloning on a large offline dataset with over 10M trajectories generated by cuRobo~\cite{curobo}, a state-of-the-art optimization-based motion planner. While this pretraining demonstrates strong global planning potential, the resulting policy often incurs minor collisions, as the kinematic expert trajectories neglect robot dynamics.

Subsequently, we enhance IMPACT's static obstacle avoidance using student-teacher finetuning. The teacher combines the pretrained IMPACT policy with Geometric Fabrics~\cite{fabrics}, a state-based closed-loop controller that excels at local obstacle avoidance while respecting robot dynamics. Since Geometric Fabrics relies on privileged obstacle information, we distill its behavior into a fine-tuned IMPACT policy that operates directly on point-cloud inputs.

To further boost reactive performance to dynamic obstacles during deployment, DRP utilizes a locally-reactive goal proposal module, DCP-RMP, that supplies real-time obstacle avoidance targets to the fine-tuned IMPACT policy.

\subsection{Large-Scale Motion Pretraining}
\label{sec:impact}
To enable supervised pretraining of a motion policy that can learn general collision-free behavior, we generate diverse and complex training scenes paired with expert trajectory solutions. While we largely follow the data generation pipeline introduced in \cite{nmp}, we replace AIT* with cuRobo as the expert motion planner. Due to its GPU acceleration, cuRobo allows us to scale data generation to 10 million expert trajectories.

We also introduce challenging scenarios where the goal itself is obstructed by the environment and thus physically unreachable. In these cases, we modify the expert trajectory to stop the robot as close as possible to the goal without colliding with the blocking obstacle. This scenario is critical to include, as in dynamic settings, obstacles like humans may temporarily block the target, and the robot must learn to avoid collisions even when the goal cannot be immediately reached.

We then train on this data using \textbf{IMPACT} (Imitating Motion Planning with Action-Chunking Transformer), a transformer-based neural motion policy architecture. IMPACT outputs joint position targets, conditioned on the obstacle point cloud $P_s \in \mathbb{R}^{N_s \times 3}$, robot point cloud $P_r \in \mathbb{R}^{N_r \times 3}$, current joint configuration $q_c \in \mathbb{R}^7$, and goal joint configuration $q_{mg} \in \mathbb{R}^7$.

During training, point cloud inputs are generated by uniformly sampling points from ground-truth mesh surfaces in simulation. During real-world deployment, scene point clouds are captured using calibrated depth cameras. We also replace points near the robot in the captured point cloud with points sampled from its mesh model, using the current joint configuration to ensure an accurate representation of its state.

To reduce computational complexity and enable real-time inference, we use set abstraction from PointNet++~\cite{pointnet++} to downsample the point clouds and generate a smaller set of latent tokens. Specifically, the scene and robot point clouds are converted into tokens $z_s \in \mathbb{R}^{K_s \times H}$ and $z_r \in \mathbb{R}^{K_r \times H}$, where $K_s < N_s$ and $K_r < N_r$. The current and target joint angles are encoded using MLPs to produce $z_c, z_{mg} \in \mathbb{R}^H$, respectively. Each input is paired with a learnable embedding: $e_s, e_r, e_c, e_{mg} \in \mathbb{R}^H$, which are added to the corresponding tokens to form the encoder input. 

The decoder processes $S$ learnable action tokens $A \in \mathbb{R}^{S \times H}$, using the encoder output as memory. The decoder outputs a sequence of $S$ delta joint actions $[\bar{q}_1, \bar{q}_2, \dots, \bar{q}_S] \in \mathbb{R}^{S \times 7}$, which are supervised using a Mean Squared Error (MSE) loss against the ground-truth actions $[q_1, q_2, \dots, q_S]$:
\[
\mathcal{L}_{BC} = \frac{1}{S} \sum_{i=1}^{S} ||q_i - \bar{q}_i||_2
\]
These delta actions are then converted into absolute joint targets and sent to the robot's low-level controller for real-time execution.

\subsection{Iterative Student-Teacher Finetuning}
\label{sec:finetune}

\begin{wrapfigure}{t}{0.62\textwidth}
\centering
\includegraphics[width=\linewidth]{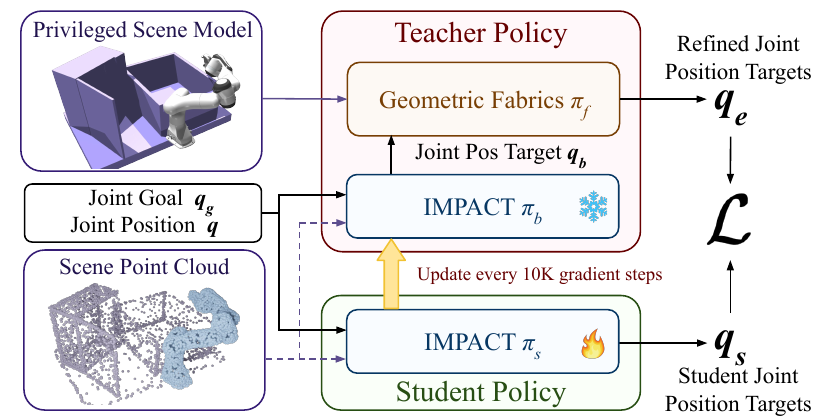}
\vspace{-0.1in}
\caption{The IMPACT policy is combined with a locally reactive Geometric Fabrics controller to enable improved obstacle avoidance.  This combined teacher policy is then distilled into a point-cloud conditioned student policy.}
\label{fig:student_teacher}
\vspace{-0.2in}
\end{wrapfigure}

Pretraining IMPACT on our dataset provides a strong globally-aware motion policy, already outperforming prior state-of-the-art neural methods (see Section~\ref{sec:sim_results}). However, since behavior cloning suffers from compounding error over long horizons, the pre-trained IMPACT policy alone still incurs frequent minor collisions at deployment.

To enhance the policy's ability for static obstacle avoidance, we improve IMPACT via iterative student-teacher finetuning. Since many of the pretrained policy's failure cases stem from minor collisions with local obstacles, we can remedy these mistakes by passing IMPACT's joint position outputs into Geometric Fabrics \cite{fabrics}, a state-of-the-art controller that excels at local obstacle avoidance.

Geometric Fabrics uses ground-truth obstacle models to follow joint targets while avoiding nearby obstacles and respecting dynamic constraints like joint jerk and acceleration limits. Since it relies on privileged information, we apply student-teacher distillation in simulation to refine our point-cloud-based IMPACT policy \cite{dagger}.

We initialize the student policy $\pi_s$ with the pretrained IMPACT policy from Section \ref{sec:impact}. The teacher policy also starts with the pretrained IMPACT policy $\pi_b$, which outputs an action chunk of joint position targets. We take the first action in this chunk, $q_b$, as an intermediate goal and pass it to Geometric Fabrics $\pi_f$, which refines it into improved targets $q_e = \pi_f(obs, q_b)$. These refined targets then supervise updates to the student policy $\pi_s$. To ensure scalability, the entire process runs in parallel using IsaacGym~\cite{isaacgym}. Since Geometric Fabrics requires signed distance fields (SDFs) for obstacle avoidance, we precompute them offline in batch for static scenes. Notably, we avoid using cuRobo as the expert during finetuning, as it would require planning at every simulation step across all vectorized environments—rendering it computationally impractical.

During distillation, we keep $\pi_b$ frozen within the teacher policy to maintain stable objectives. After 10000 gradient steps, the fine-tuned student replaces $\pi_b$, and the process repeats. This iterative procedure progressively improves local obstacle avoidance while preserving strong global planning capabilities. The full process is illustrated in Figure~\ref{fig:student_teacher}. This iterative student-teacher finetuning improves the success rate over the pretrained model by 45\%, as shown in Table~\ref{tab:sim_results}.

\begin{figure}[t]
    \centering
\includegraphics[width=0.99\textwidth]{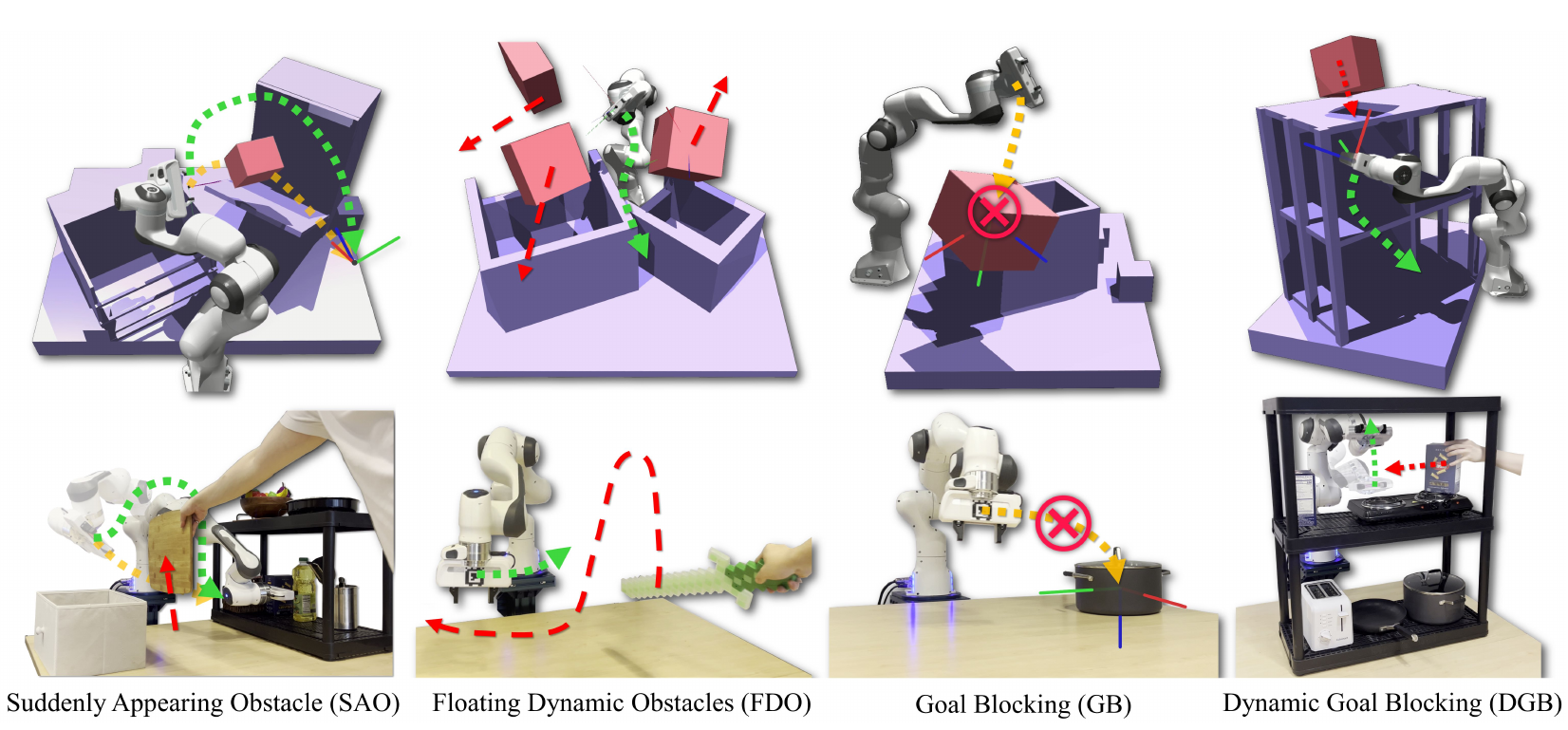}
    \caption{\textbf{DRPBench} introduces five challenging evaluation scenarios in simulation and real. In addition to complex Static Environments \textbf{(SE)}, we propose Suddenly Appearing Obstacle \textbf{(SAO)} where obstacles appear suddenly ahead of the robot, Floating Dynamic Obstacles \textbf{(FDO)} where obstacles move randomly throughout the environment, Goal Blocking \textbf{(GB)} where the goal is obstructed and the robot must approach as closely as possible without colliding, and Dynamic Goal Blocking \textbf{(DGB)} where the robot encounters a moving obstacle \textit{after} getting to the goal.}\label{fig:exp}
\vspace{-1.7em}
\end{figure}

\subsection{Riemannian Motion Policy with Dynamic Closest Point}
\vspace{-0.05in}
\label{sec:rmp}
While IMPACT's closed-loop nature allows it to implicitly handle dynamic environments—making decisions at every timestep—its performance degrades in particularly challenging scenarios, such as when an object moves rapidly toward the robot. This limitation arises because dynamic interactions are absent from both the pretraining dataset and the student-teacher finetuning phase, as generating expert trajectories for non-static scenes would require re-planning after every action, which is computationally infeasible for both cuRobo and our vectorized Geometric Fabrics implementation.

To explicitly enhance IMPACT's dynamic obstacle avoidance during inference, we introduce a Riemannian Motion Policy (RMP) layer. This non-learning based component uses local obstacle information to enable highly reactive local obstacle avoidance. Specifically, RMP acts as a goal-proposal module, modifying the original joint-space goal $\mathbf{q}_g$ into a new goal $\mathbf{q}_{mg}$ that prioritizes avoidance when dynamic obstacles approach. This adjusted goal is then passed to IMPACT. Notably, because IMPACT is already trained with global scene awareness, focusing RMP solely on local dynamic obstacles does not compromise the global goal-reaching performance.

However, RMP traditionally requires ground-truth obstacle models and poses to generate joint targets for reactive avoidance, limiting its real-world deployability. To overcome this, we propose Dynamic Closest Point RMP (DCP-RMP)—an RMP variant that operates directly on point cloud inputs. At a high level, DCP-RMP identifies the closest point in the point cloud belonging to a dynamic obstacle and generates repulsive motion to steer the robot away.

Specifically, we implement DCP-RMP by first extracting the dynamic obstacle point cloud using a KDTree, which efficiently performs nearest-neighbor queries between the current and previous frame point clouds to identify moving points. We then compute the minimal displacement $\mathbf{x}_r$ between the robot and nearby dynamic obstacles and derive a repulsive acceleration to increase this separation. Finally, we adjust the original joint goal $\mathbf{q}_d$ by virtually applying this repulsive signal, yielding the modified goal $\mathbf{q}_{mg}$ that prioritizes dynamic obstacle avoidance. Detailed mathematical formulations are provided in Appendix~\ref{sec:RMP_details}.

While the modified joint goal may sometimes intersect with static obstacles, rendering it physically unachievable, IMPACT has been trained on scenarios where the goal configuration is in collision with the scene and learns to stop safely in front of obstacles instead.


\begin{table}[t]
    \centering
    \setlength{\tabcolsep}{10pt}
    \begin{tabular}{@{}lccccc@{\hspace{1pt}}c@{}}
    \toprule
    & \multicolumn{5}{c}{\textbf{DRPBench}} & \multirow{2}{*}{\textbf{M$\pi$Nets Dataset}} \\
    \cmidrule(lr){2-6}
    & SE & SAO & FDO & GB & DGB & \\
    \midrule
    \multicolumn{7}{@{}l}{\textit{With privileged information}:} \\
    AIT* & 40.50 & 0 & 0 & 0 & 0 & 89.22 \\
    cuRobo & 82.97 & 59.00 & 39.50 & 0 & 3.00 & \textbf{99.78} \\
    \midrule
    cuRobo-Vox & 50.53 & 58.67 & 40.50 & 0 & 2.50 & - \\
    RMP & 32.97 & 46.0 & 49.50 & \textbf{71.08} & 50.50 & 41.90 \\
    \midrule
    M$\pi$Nets & 2.50 & 0.33 & 0 & 0 & 0 & 65.18 \\
    M$\pi$Former & 0.19 & 0 & 0 & 0 & 0 & 30.02 \\
    NeuralMP & 50.59 & 33.16 & 19.00 & 0 & 0.25 & -- \\
    \midrule
    IMPACT (no finetune) & 58.25 & 47.33 & 13.50 & 46.50 & 0 & 66.27 \\
    IMPACT & \textbf{84.60} & \textbf{86.00} & 32.00 & 66.67 & 0.25 & 83.71 \\
    \textit{DRP (Ours)} & \textbf{84.60} & \textbf{86.00} & \textbf{75.50} & 66.67 & \textbf{65.25} & 83.71 \\
    \bottomrule
    \end{tabular}
    \vspace{1em}
    \caption{\small Quantitative results on DRPBench and M$\pi$Nets Dataset. DRP outperforms all classical and learning-based baselines across diverse settings, particularly excelling in dynamic and goal-blocking tasks. While optimization methods like cuRobo succeed in static scenes, they struggle under dynamic conditions. DRP's combination of architectural improvements, fine-tuning, and reactive control (via DCP-RMP) enables robust generalization and superior performance, especially in scenarios requiring fast adaptation.}\label{tab:sim_results}
\vspace{-2.2em}
\end{table}

\section{Experiment Setup and Results}\label{sec:exp}

To comprehensively evaluate DRP's reactivity and robustness, we design \textbf{DRPBench}, a set of challenging benchmark tasks in both simulation and the real world. These tasks target three critical real-world challenges for robot motion generation: navigating cluttered static environments, reacting swiftly to dynamic obstacles, and handling temporarily obstructed goals with caution and precision.

We report quantitative results and address the following key questions: (1) How does DRP perform compared to state-of-the-art classical and learning-based motion planners? (2) What are the individual contributions of DRP's architectural design, student-teacher fine-tuning, and DCP-RMP integration? (3) Can DRP generalize effectively to real-world environments, especially those exhibiting significant domain shift from training?

\subsection{Simulation Experiments and Results}\label{sec:sim_results}

For the simulation experiments, \textbf{DRPBench} comprises over 4000 diverse problem instances, with examples shown in Figure~\ref{fig:exp}. Further details about tasks are provided in the Appendix. In addition to DRPBench, we also test on the M$\pi$Nets benchmark~\cite{mpinet} in a zero-shot manner without additional training.  We use success rate as the primary metric, where a trial is successful if the robot reaches the end-effector goal pose within position and orientation thresholds without collisions. 

\textbf{DRP excels in dynamic and goal-blocking scenes.}
Sampling-based planners such as AIT* completely fail in dynamic environments, achieving 0\% success on all such tasks despite extended planning horizons. Optimization-based approaches like cuRobo and RMP perform significantly better in static settings—e.g., 82.97\% and 32.97\% on Static Environment (SE), respectively—but degrade in harder scenarios. cuRobo drops to 3.00\% on Dynamic Goal Blocking (DGB), where the goal is temporarily obstructed. In contrast, RMP achieves 50.50\% on DGB, motivating its integration into DRP as a reactive control module. Still, DRP surpasses RMP significantly on tasks except for Goal Blocking (GB), underscoring the benefit of combining learning with reactive control.

\textbf{Architectural design and training diversity enables generalization.}
Learning-based models trained on narrow datasets—such as M$\pi$Nets and M$\pi$Former—fail to generalize to out-of-distribution scenes, achieving only 0–2.5\% on the DRPBench tasks. NeuralMP performs better, but it relies on test-time optimization (TTO)—a process that adapts trajectories post-hoc during deployment which only helps in Static Environment (SE) and struggles in reactive contexts. However, even without finetuning, IMPACT outperforms other learning-based methods due to its more expressive, scalable architecture and training on a more diverse dataset—without relying on post-hoc techniques. We additionally include architecture ablations, demonstrating that IMPACT outperforms other architectures from prior learning-based methods while training on the same data; see Appendix~\ref{arch-ablation}.

\textbf{Fine-tuning surpasses data generation method.}
DRP is pretrained using trajectories from a classical planner (cuRobo), but it significantly exceeds this source's performance. This is because we filter for successfully planned trajectories during data generation and we apply a student-teacher fine-tuning stage that distills and refines action generation beyond the original data. The result is a policy that not only inherits useful behaviors but generalizes more effectively across complex settings. IMPACT also performs well in static and dynamic environments that require fine local control. It achieves 86.00\% on Suddenly Appearing Obstacle (SAO) and 66.67\% on Goal Blocking (GB), where precise maneuvers near occluded or blocked targets are critical. These results validate the strength of closed-loop visuo-motor imitation in spatially constrained environments.

\textbf{DCP-RMP boosts dynamic performance.}
DRP's integration of DCP-RMP adds fast local responsiveness, yielding strong results in fully dynamic tasks: 75.50\% on FDO and 65.25\% on DGB. In contrast, using IMPACT alone drops to 32.00\% and 0.25\% on the same tasks, while cuRobo reaches only 39.50\% and 3.00\%. These gains highlight the necessity of combining high-level learning with reactive modules to handle dynamic obstacles and shifting goals in real time. We further evaluate the impact of the DCP-RMP module by adding it to various baseline methods. As shown in Table~\ref{tab:ablate-rmp}, DCP-RMP improves performance across all dynamic tasks, regardless of the underlying method.

\begin{table}[t]
\centering
\renewcommand{\arraystretch}{1.1}
\begin{minipage}[t]{0.48\textwidth}
\centering
\setlength{\tabcolsep}{4pt}
\begin{tabular}{lcccc}
    \toprule
    & \multicolumn{2}{c}{FDO} & \multicolumn{2}{c}{DGB} \\
    \cmidrule(lr){2-3} \cmidrule(lr){4-5}
    Has DCP-RMP? & \ding{55} & \ding{51} & \ding{55} & \ding{51} \\
    \midrule
    cuRobo~\cite{curobo} & 39.50 & 51.50 & 3.00 & 54.50 \\
    NeuralMP~\cite{nmp} & 19.00 & 34.00 & 0.25 & 20.75 \\
    IMPACT & 32.00 & \textbf{75.50} & 0.25 & \textbf{65.25} \\
    \bottomrule
\end{tabular}
\vspace{2pt}
\caption{\small DCP-RMP is method-agnostic.}\label{tab:ablate-rmp}
\end{minipage}
\vspace{-1.5em}
\hfill
\begin{minipage}[t]{0.48\textwidth}
\centering
\setlength{\tabcolsep}{1.5pt} 
\begin{tabular}{@{}lccccc@{}}
    \toprule
    & SE & SAO & FDO & GB & DGB \\
    \midrule
    cuRobo-Vox & 60.00 & 3.33 & 0 & 0 & 0 \\
    NeuralMP~\cite{nmp} & 30.00 & 6.67 & 0 & 0 & 0 \\
    IMPACT & \textbf{90.00} & \textbf{100.00} & 0 & \textbf{92.86} & 0 \\
    \textit{DRP (Ours)} & \textbf{90.00} & \textbf{100.00} & \textbf{70.00} & \textbf{92.86} & \textbf{93.33} \\
    \bottomrule
\end{tabular}
\vspace{2pt}
\caption{\small Success Rate (\%) on Real-world tasks.}\label{tab:real_results}
\end{minipage}
\vspace{-1.5em}
\end{table}

\subsection{Real-World Experiment Results}\label{sec:real_world}
\vspace{-0.05in}

Our real-world benchmark mirrors the five simulation tasks, but introduces out-of-distribution, semantically-meaningful obstacles like a slanted shelf and tall drawer, as shown in Figure~\ref{fig:teaser}. For example, in one dynamic goal blocking (DGB) task, the robot must wait in front of a drawer, avoid a human operator, and reach in once it opens. The benchmark includes over 50 real-world instances, with two to four calibrated Intel RealSense D455 RGB-D cameras capturing the scene.

As seen in Table~\ref{tab:real_results}, DRP significantly outperforms classical and learning-based baselines in real-world settings. On tasks like Static Environment (SE) and Static Appearing Obstruction (SAO), both DRP and IMPACT achieve near-perfect success rates, far exceeding cuRobo-Vox and NeuralMP, which degrade severely under noisy perception. On the more challenging Goal Blocking (GB) task, both DRP and IMPACT reach 92.86\%, while cuRobo and NeuralMP fail completely. The biggest difference appears on Floating Dynamic Obstacle (FDO) and Dynamic Goal Blocking (DGB) where IMPACT fails to solve, highlighting the value of DCP-RMP for reactive behavior. Even though  being trained entirely in simulation, DRP adapts well to real-world settings.

\vspace{-0.05in}
\section{Conclusion}
\vspace{-0.05in}
\label{sec:conclusion}

We introduced Deep Reactive Policy (DRP), a scalable, generalizable framework for closed-loop motion generation in complex and dynamic environments. At its core is IMPACT, a transformer-based visuo-motor policy trained on large-scale motion data and refined via student-teacher finetuning.  DRP learns directly from point cloud inputs to produce collision-free, globally coherent actions while remaining robust to partial observability and environmental changes. Extensive evaluations in both simulation and real-world settings show DRP consistently outperforms prior learning-based and classical planners, especially in scenarios requiring reactive adaptation. While IMPACT addresses most planning challenges end-to-end, incorporating lightweight reactive modules like DCP-RMP further boosts performance in highly dynamic scenes. To support ongoing research, we will release all datasets, models, and benchmarks.

\section{Limitations}
\label{sec:limitations}
DRP relies on reasonably accurate point cloud observations for effective planning. Although the policy is robust to a certain degree of noise and partial observability, performance may degrade under severe perception failures. Our multi-camera setup helps mitigate this issue, but may not be suffice for tasks in narrow environments with frequent occlusions.  Leveraging RGB or RGB-D inputs could improve performance for more unstructured environments.

Our experiments are limited to a single embodiment—the Franka Panda—and we do not evaluate on other robot platforms. This limitation stems from the challenges in scaling our current pipeline to multiple embodiments. In future work, we aim to address this by either generating separate planners for each robot or training a single DRP policy that generalizes across embodiments.

\acknowledgments{}
We thank Murtaza Dalal, Ritvik Singh, Arthur Allshire, Tal Daniel, Zheyuan Hu, Mohan Kumar Srirama, and Ruslan Salakhutdinov for their valuable discussions on this work. We are grateful to Karl Van Wyk and Nathan Ratliff for contributing ideas and implementations of Geometric Fabrics used in this project. We also thank Murtaza Dalal for his feedback on the early ideations of this paper. In addition, we thank Andrew Wang, Tony Tao, Hengkai Pan, Tiffany Tse, Sheqi Zhang, and Sungjae Park for their assistance with experiments. This work is supported in part by ONR MURI N00014-22-1-2773, ONR MURI N00014-24-1-2748, and AFOSR FA9550-23-1-0747. 

\clearpage


\bibliography{references}  

\begin{thebibliography}{35}
\providecommand{\natexlab}[1]{#1}
\providecommand{\url}[1]{\texttt{#1}}
\expandafter\ifx\csname urlstyle\endcsname\relax
  \providecommand{\doi}[1]{doi: #1}\else
  \providecommand{\doi}{doi: \begingroup \urlstyle{rm}\Url}\fi

\bibitem[Hart et~al.(1968)Hart, Nilsson, and Raphael]{a*}
P.~E. Hart, N.~J. Nilsson, and B.~Raphael.
\newblock A formal basis for the heuristic determination of minimum cost paths.
\newblock \emph{IEEE transactions on Systems Science and Cybernetics}, 4\penalty0 (2):\penalty0 100--107, 1968.

\bibitem[Strub and Gammell(2020)]{ait*}
M.~P. Strub and J.~D. Gammell.
\newblock Adaptively informed trees (ait*): Fast asymptotically optimal path planning through adaptive heuristics.
\newblock In \emph{2020 IEEE International Conference on Robotics and Automation (ICRA)}, pages 3191--3198. IEEE, 2020.

\bibitem[Khatib(1985)]{khatib1985real}
O.~Khatib.
\newblock Real-time obstacle avoidance for manipulators and mobile robots.
\newblock In \emph{Proceedings. 1985 IEEE international conference on robotics and automation}, volume~2, pages 500--505. IEEE, 1985.

\bibitem[Ratliff et~al.(2018)Ratliff, Issac, Kappler, Birchfield, and Fox]{rmp}
N.~D. Ratliff, J.~Issac, D.~Kappler, S.~Birchfield, and D.~Fox.
\newblock Riemannian motion policies.
\newblock \emph{arXiv preprint arXiv:1801.02854}, 2018.

\bibitem[Van~Wyk et~al.(2022)Van~Wyk, Xie, Li, Rana, Babich, Peele, Wan, Akinola, Sundaralingam, Fox, et~al.]{fabrics}
K.~Van~Wyk, M.~Xie, A.~Li, M.~A. Rana, B.~Babich, B.~Peele, Q.~Wan, I.~Akinola, B.~Sundaralingam, D.~Fox, et~al.
\newblock Geometric fabrics: Generalizing classical mechanics to capture the physics of behavior.
\newblock \emph{IEEE Robotics and Automation Letters}, 7\penalty0 (2):\penalty0 3202--3209, 2022.

\bibitem[Bhardwaj et~al.(2022)Bhardwaj, Sundaralingam, Mousavian, Ratliff, Fox, Ramos, and Boots]{bhardwaj2022storm}
M.~Bhardwaj, B.~Sundaralingam, A.~Mousavian, N.~D. Ratliff, D.~Fox, F.~Ramos, and B.~Boots.
\newblock Storm: An integrated framework for fast joint-space model-predictive control for reactive manipulation.
\newblock In \emph{Conference on Robot Learning}, pages 750--759. PMLR, 2022.

\bibitem[Fishman et~al.(2023)Fishman, Murali, Eppner, Peele, Boots, and Fox]{mpinet}
A.~Fishman, A.~Murali, C.~Eppner, B.~Peele, B.~Boots, and D.~Fox.
\newblock Motion policy networks.
\newblock In \emph{conference on Robot Learning}, pages 967--977. PMLR, 2023.

\bibitem[Qureshi et~al.(2019)Qureshi, Simeonov, Bency, and Yip]{mpnet}
A.~H. Qureshi, A.~Simeonov, M.~J. Bency, and M.~C. Yip.
\newblock Motion planning networks.
\newblock In \emph{2019 International Conference on Robotics and Automation (ICRA)}, pages 2118--2124. IEEE, 2019.

\bibitem[Dalal et~al.(2024)Dalal, Yang, Mendonca, Khaky, Salakhutdinov, and Pathak]{nmp}
M.~Dalal, J.~Yang, R.~Mendonca, Y.~Khaky, R.~Salakhutdinov, and D.~Pathak.
\newblock Neural mp: A generalist neural motion planner.
\newblock \emph{arXiv preprint arXiv:2409.05864}, 2024.

\bibitem[Fishman et~al.(2024)Fishman, Walsman, Bhardwaj, Yuan, Sundaralingam, Boots, and Fox]{mpiformer}
A.~Fishman, A.~Walsman, M.~Bhardwaj, W.~Yuan, B.~Sundaralingam, B.~Boots, and D.~Fox.
\newblock Avoid everything: Model-free collision avoidance with expert-guided fine-tuning.
\newblock In \emph{CoRL Workshop on Safe and Robust Robot Learning for Operation in the Real World}, 2024.

\bibitem[Sundaralingam et~al.(2023)Sundaralingam, Hari, Fishman, Garrett, Van~Wyk, Blukis, Millane, Oleynikova, Handa, Ramos, et~al.]{curobo}
B.~Sundaralingam, S.~K.~S. Hari, A.~Fishman, C.~Garrett, K.~Van~Wyk, V.~Blukis, A.~Millane, H.~Oleynikova, A.~Handa, F.~Ramos, et~al.
\newblock Curobo: Parallelized collision-free robot motion generation.
\newblock In \emph{2023 IEEE International Conference on Robotics and Automation (ICRA)}, pages 8112--8119. IEEE, 2023.

\bibitem[Likhachev et~al.(2003)Likhachev, Gordon, and Thrun]{likhachev2003ara}
M.~Likhachev, G.~J. Gordon, and S.~Thrun.
\newblock Ara*: Anytime a* with provable bounds on sub-optimality.
\newblock \emph{Advances in neural information processing systems}, 16, 2003.

\bibitem[Likhachev et~al.(2005)Likhachev, Ferguson, Gordon, Stentz, and Thrun]{likhachev2005anytime}
M.~Likhachev, D.~I. Ferguson, G.~J. Gordon, A.~Stentz, and S.~Thrun.
\newblock Anytime dynamic a*: An anytime, replanning algorithm.
\newblock In \emph{ICAPS}, volume~5, pages 262--271, 2005.

\bibitem[Koenig and Likhachev(2006)]{koenig2006new}
S.~Koenig and M.~Likhachev.
\newblock A new principle for incremental heuristic search: Theoretical results.
\newblock In \emph{ICAPS}, pages 402--405, 2006.

\bibitem[Kavraki et~al.(1996)Kavraki, Svestka, Latombe, and Overmars]{prm}
L.~E. Kavraki, P.~Svestka, J.-C. Latombe, and M.~H. Overmars.
\newblock Probabilistic roadmaps for path planning in high-dimensional configuration spaces.
\newblock \emph{IEEE transactions on Robotics and Automation}, 12\penalty0 (4):\penalty0 566--580, 1996.

\bibitem[LaValle(1998)]{rrt}
S.~LaValle.
\newblock Rapidly-exploring random trees: A new tool for path planning.
\newblock \emph{Research Report 9811}, 1998.

\bibitem[Bohlin and Kavraki(2000)]{lazyprm}
R.~Bohlin and L.~E. Kavraki.
\newblock Path planning using lazy prm.
\newblock In \emph{Proceedings 2000 ICRA. Millennium conference. IEEE international conference on robotics and automation. Symposia proceedings (Cat. No. 00CH37065)}, volume~1, pages 521--528. IEEE, 2000.

\bibitem[LaValle and Kuffner(2001)]{lavalle2001rapidly}
S.~M. LaValle and J.~J. Kuffner.
\newblock Rapidly-exploring random trees: Progress and prospects: Steven m. lavalle, iowa state university, a james j. kuffner, jr., university of tokyo, tokyo, japan.
\newblock \emph{Algorithmic and computational robotics}, pages 303--307, 2001.

\bibitem[Kuffner and LaValle(2000)]{kuffner2000rrt}
J.~J. Kuffner and S.~M. LaValle.
\newblock Rrt-connect: An efficient approach to single-query path planning.
\newblock In \emph{Proceedings 2000 ICRA. Millennium Conference. IEEE International Conference on Robotics and Automation. Symposia Proceedings (Cat. No. 00CH37065)}, volume~2, pages 995--1001. IEEE, 2000.

\bibitem[Gammell et~al.(2015)Gammell, Srinivasa, and Barfoot]{bit*}
J.~D. Gammell, S.~S. Srinivasa, and T.~D. Barfoot.
\newblock Batch informed trees (bit*): Sampling-based optimal planning via the heuristically guided search of implicit random geometric graphs.
\newblock In \emph{2015 IEEE International Conference on Robotics and Automation (ICRA)}, pages 3067--3074, 2015.
\newblock \doi{10.1109/ICRA.2015.7139620}.

\bibitem[Zucker et~al.(2013)Zucker, Ratliff, Dragan, Pivtoraiko, Klingensmith, Dellin, Bagnell, and Srinivasa]{zucker2013chomp}
M.~Zucker, N.~Ratliff, A.~D. Dragan, M.~Pivtoraiko, M.~Klingensmith, C.~M. Dellin, J.~A. Bagnell, and S.~S. Srinivasa.
\newblock Chomp: Covariant hamiltonian optimization for motion planning.
\newblock \emph{The International journal of robotics research}, 32\penalty0 (9-10):\penalty0 1164--1193, 2013.

\bibitem[Schulman et~al.(2014)Schulman, Duan, Ho, Lee, Awwal, Bradlow, Pan, Patil, Goldberg, and Abbeel]{schulman2014motion}
J.~Schulman, Y.~Duan, J.~Ho, A.~Lee, I.~Awwal, H.~Bradlow, J.~Pan, S.~Patil, K.~Goldberg, and P.~Abbeel.
\newblock Motion planning with sequential convex optimization and convex collision checking.
\newblock \emph{The International Journal of Robotics Research}, 33\penalty0 (9):\penalty0 1251--1270, 2014.

\bibitem[Dragan et~al.(2011)Dragan, Ratliff, and Srinivasa]{dragan2011manipulation}
A.~D. Dragan, N.~D. Ratliff, and S.~S. Srinivasa.
\newblock Manipulation planning with goal sets using constrained trajectory optimization.
\newblock In \emph{2011 IEEE International Conference on Robotics and Automation}, pages 4582--4588. IEEE, 2011.

\bibitem[Kumar et~al.(2019)Kumar, Mandalika, Choudhury, and Srinivasa]{kumar2019lego}
R.~Kumar, A.~Mandalika, S.~Choudhury, and S.~Srinivasa.
\newblock Lego: Leveraging experience in roadmap generation for sampling-based planning.
\newblock In \emph{2019 IEEE/RSJ International Conference on Intelligent Robots and Systems (IROS)}, pages 1488--1495. IEEE, 2019.

\bibitem[Zhang et~al.(2018)Zhang, Huh, and Lee]{zhanglearning}
C.~Zhang, J.~Huh, and D.~D. Lee.
\newblock Learning implicit sampling distributions for motion planning. in 2018 ieee.
\newblock In \emph{RSJ International Conference on Intelligent Robots and Systems (IROS)}, pages 3654--3661, 2018.

\bibitem[Chamzas et~al.(2021)Chamzas, Kingston, Quintero-Pe{\~n}a, Shrivastava, and Kavraki]{chamzas2021learning}
C.~Chamzas, Z.~Kingston, C.~Quintero-Pe{\~n}a, A.~Shrivastava, and L.~E. Kavraki.
\newblock Learning sampling distributions using local 3d workspace decompositions for motion planning in high dimensions.
\newblock In \emph{2021 IEEE International Conference on Robotics and Automation (ICRA)}, pages 1283--1289. IEEE, 2021.

\bibitem[Ichter et~al.(2020)Ichter, Sermanet, and Lynch]{belt}
B.~Ichter, P.~Sermanet, and C.~Lynch.
\newblock Broadly-exploring, local-policy trees for long-horizon task planning.
\newblock \emph{arXiv preprint arXiv:2010.06491}, 2020.

\bibitem[Qureshi et~al.(2020)Qureshi, Dong, Choe, and Yip]{compnet}
A.~H. Qureshi, J.~Dong, A.~Choe, and M.~C. Yip.
\newblock Neural manipulation planning on constraint manifolds.
\newblock \emph{IEEE Robotics and Automation Letters}, 5\penalty0 (4):\penalty0 6089--6096, 2020.

\bibitem[Huang et~al.(2024)Huang, Sundaralingam, Mousavian, Murali, Goldberg, and Fox]{diffusionseeder}
H.~Huang, B.~Sundaralingam, A.~Mousavian, A.~Murali, K.~Goldberg, and D.~Fox.
\newblock Diffusionseeder: Seeding motion optimization with diffusion for rapid motion planning.
\newblock \emph{arXiv preprint arXiv:2410.16727}, 2024.

\bibitem[Johnson et~al.(2020)Johnson, Li, Liu, Qureshi, and Yip]{dynmpnet}
J.~J. Johnson, L.~Li, F.~Liu, A.~H. Qureshi, and M.~C. Yip.
\newblock Dynamically constrained motion planning networks for non-holonomic robots.
\newblock In \emph{2020 IEEE/RSJ International Conference on Intelligent Robots and Systems (IROS)}, pages 6937--6943. IEEE, 2020.

\bibitem[Carvalho et~al.(2023)Carvalho, Le, Baierl, Koert, and Peters]{mpdiffusion}
J.~Carvalho, A.~T. Le, M.~Baierl, D.~Koert, and J.~Peters.
\newblock Motion planning diffusion: Learning and planning of robot motions with diffusion models.
\newblock In \emph{2023 IEEE/RSJ International Conference on Intelligent Robots and Systems (IROS)}, pages 1916--1923. IEEE, 2023.

\bibitem[Saha et~al.(2024)Saha, Mandadi, Reddy, Srikanth, Agarwal, Sen, Singh, and Krishna]{edmp}
K.~Saha, V.~Mandadi, J.~Reddy, A.~Srikanth, A.~Agarwal, B.~Sen, A.~Singh, and M.~Krishna.
\newblock Edmp: Ensemble-of-costs-guided diffusion for motion planning.
\newblock In \emph{2024 IEEE International Conference on Robotics and Automation (ICRA)}, pages 10351--10358. IEEE, 2024.

\bibitem[Qi et~al.(2017)Qi, Yi, Su, and Guibas]{pointnet++}
C.~R. Qi, L.~Yi, H.~Su, and L.~J. Guibas.
\newblock Pointnet++: Deep hierarchical feature learning on point sets in a metric space.
\newblock \emph{Advances in neural information processing systems}, 30, 2017.

\bibitem[Ross et~al.(2011)Ross, Gordon, and Bagnell]{dagger}
S.~Ross, G.~Gordon, and D.~Bagnell.
\newblock A reduction of imitation learning and structured prediction to no-regret online learning.
\newblock In \emph{Proceedings of the fourteenth international conference on artificial intelligence and statistics}, pages 627--635. JMLR Workshop and Conference Proceedings, 2011.

\bibitem[Makoviychuk et~al.(2021)Makoviychuk, Wawrzyniak, Guo, Lu, Storey, Macklin, Hoeller, Rudin, Allshire, Handa, et~al.]{isaacgym}
V.~Makoviychuk, L.~Wawrzyniak, Y.~Guo, M.~Lu, K.~Storey, M.~Macklin, D.~Hoeller, N.~Rudin, A.~Allshire, A.~Handa, et~al.
\newblock Isaac gym: High performance gpu-based physics simulation for robot learning.
\newblock \emph{arXiv preprint arXiv:2108.10470}, 2021.

\end{thebibliography}

\clearpage

\appendix
\part*{Appendix}
\section{Detailed Task Descriptions}

We propose \textbf{DRPBench} in simulation which comprises of five tasks:
\vspace{-0.5em}
\begin{itemize}[leftmargin=1.5em]
    \item \textbf{Static Environments (SE):} These scenarios feature challenging fixed obstacles, evaluating policies performance in predictable, unchanging settings.
    \item \textbf{Suddenly Appearing Obstacle (SAO):} Obstacles appear suddenly ahead of the robot, directly blocking its path and requiring dynamic trajectory adaptation. This tests the policy's ability to react to unexpected changes in the environment.
    \item \textbf{Floating Dynamic Obstacles (FDO):} Obstacles move randomly throughout the environment, challenging the robot's reactive capabilities and its ability to avoid collisions in real time.
    \item \textbf{Goal Blocking (GB):} 
    The goal is temporarily obstructed by an obstacle, and the robot must approach as closely as possible without colliding.
    \item \textbf{Dynamic Goal Blocking (DGB):} After reaching the goal, the robot encounters a moving obstacle and must avoid it before safely returning to the goal, testing its ability to remain reactive even after task completion.
\end{itemize}
\vspace{-0.5em}
We use the same success metric as prior works \cite{nmp, mpinet}, where a goal is deemed reached if the end-effector translation error is within $1$ cm and orientaion error is within $ 15^{\circ}$.

\section{Baseline Comparison Details}

DRP is compared against a broad set of baselines, including both classical planning methods and learning-based approaches. For \textit{classical methods}, we consider the following:

\vspace{-0.5em}
\begin{itemize}[leftmargin=1.5em]
    \item \textbf{AIT*}~\cite{ait*}: A state-of-the-art sampling-based planner that requires pre-computation and precise object models of the obstacles. We enforce an 80-second limit for this pre-computation.
    \item \textbf{cuRobo}~\cite{curobo}: A state-of-the-art optimization-based planner, also used as the expert policy for generating our pretraining dataset.
    \item \textbf{cuRobo-Vox}~\cite{curobo}: Instead of precise object models of the obstacles, we used voxelized inputs derived from point clouds to enable deployment in unstructured environments.
    \item \textbf{RMP}~\cite{rmp}: A state-of-the-art locally-reactive method.
\end{itemize}
\vspace{-0.5em}

For \textit{learning-based methods}, we compare against several recent works:

\vspace{-0.5em}
\begin{itemize}[leftmargin=1.5em]
    \item \textbf{M$\pi$Nets}~\cite{mpinet}: A state-of-the-art neural policy, though trained on a less diverse dataset.
    \item \textbf{M$\pi$Former}~\cite{mpiformer}: A follow-up to M$\pi$Nets that adopts a more modern architecture and incorporates a teacher-student fine-tuning stage with hard negative mining. It trains separate policies for different task scenarios; we report the best-performing checkpoint for each task.
    \item \textbf{NeuralMP}~\cite{nmp}: An end-to-end motion generation policy trained on a more diverse dataset, but requiring a pre-computation phase known as test-time optimization (TTO) to boost performance.
\end{itemize}
\vspace{-0.5em}


\section{DCP-RMP Details}
\label{sec:RMP_details}
A Riemannian Motion Policy (RMP) \cite{rmp} is a non-learning-based motion policy that takes the robot and environment state as input and outputs joint-space accelerations. RMP naturally fuses multiple single-purpose motion policies—such as moving toward a goal or repelling from an obstacle—into a single combined policy.

Each single-purpose policy is paired with a state-dependent priority metric. These policies are first transformed into a common space (typically joint space), weighted by their respective metrics, and summed together. The result is effectively a metric-weighted combination that allows selective prioritization or suppression of single-purpose policies based on the current state.

In DRP-RMP, we combine two hand-designed single-purpose policies: one for goal attraction and one for dynamic obstacle avoidance.

\textbf{Goal-Attracting Policy.} We define a simple goal-attracting acceleration policy $\mathbf{f}_g$ in joint space $\mathcal{Q}$, which pulls the robot joint position towards a desired joint position. It has an associated metric $\mathbf{M}_g$: 
\begin{equation*}
    \ddot{\mathbf{q}}_g = \mathbf{f}_g (\mathbf{q}, \dot{\mathbf{q}}) = k_{g} (\mathbf{q}_g - \mathbf{q}) - k_d \dot{\mathbf{q}} \quad \text{and} \quad \mathbf{M}_g = \mu_g \mathbf{I}
\end{equation*}

\textbf{Dynamic Obstacle-Avoiding Policy.} Traditional implementations of obstacle-avoidance policy requires ground-truth obstacle information such as obstacle models and poses. In order to deploy this policy in the real-world, we propose a point-cloud based policy. Specifically, we define the task-space of this policy $\mathcal{R}$ and its associated task-space Jacobian $\mathbf{J}_r$  with respect to the joint space $\mathcal{Q}$ as follows:
\begin{equation*}
    \mathbf{x}_r= ||\mathbf{x}_p - \mathbf{x}_{obs}|| ^2
    \quad \text{and} \quad
    \mathbf{J}_r(\mathbf{q}) = \frac{\partial \mathbf{x}_r}{\partial \mathbf{q}} =     
    2(\mathbf{x}_p - \mathbf{x}_{obs})^{\top} \mathbf{J}_p(\mathbf{q})
\end{equation*}
where $\mathbf{x}_{obs}$ is the closest dynamic obstacle point to the robot,  $\mathbf{x}_{p}$ is the closest point on the robot surface to $\mathbf{x}_{obs}$, and $\mathbf{J}_p$ is the manipulator Jacobian at robot surface point $\mathbf{x}_{p}$. To extract $\mathbf{x}_{obs}$, we build a KD-Tree from the scene point cloud captured in the previous frame and query it using the current frame's point cloud to identify points not present previously, thereby classifying them as dynamic points. From these dynamic points, we select the one closest to the robot surface. We define a dynamic obstacle-avoiding acceleration policy $\mathbf{f}_r$ and its metric as $\mathbf{M_r}$ as follows:

\begin{equation*}
    \ddot{\mathbf{x}}_r = \mathbf{f}_r ( \mathbf{x}_r, \dot{\mathbf{x}}_r) = k_p \exp(-\mathbf{x}_r / \ell_p) - k_v \left[ 1 - \frac{1}{1 + \exp(-\dot{\mathbf{x}}_r / l_v)} \right] \frac{\dot{\mathbf{x}}_r}{\mathbf{x}_r / \ell_d + \epsilon_d}
\end{equation*}
\begin{equation*}
\mathbf{M_r}(\mathbf{x}, \dot{\mathbf{x}}_r) = \left[ 1 - \frac{1}{1 + \exp(-\dot{\mathbf{x}}_r / l_v)} \right] g(\mathbf{x}_r) \frac{\mu_r}{\mathbf{x}_r / \ell_m + \epsilon_m},
\quad
g(x) = 
\begin{cases}
    (\mathbf{x}_r-r)^2/r^2, & x \leq r \\
    0, & x > r
\end{cases}
\end{equation*}

Intuitively, the obstacle-avoidance policy $\mathbf{f}_r(\mathbf{x}_r, \dot{\mathbf{x}}_r)$ generates accelerations pushing away from $\mathbf{x}_{obs}$ that grow stronger as the distance $\mathbf{x}_r$ decreases. Additionally, the repulsive acceleration increases if $\mathbf{x}_r$ is closing more rapidly. The associated metric $\mathbf{M}_r(\mathbf{x}_r, \dot{\mathbf{x}}_r)$ amplifies the influence of this policy on the combined RMP when the closing speed increases and fully deactivates the policy when the obstacle is beyond a threshold distance $r$.

Since the dynamic obstacle-avoiding policy is defined in task-space $\mathcal{R}$, we need to transform it back to joint-space $\mathcal{Q}$ to be combined with $\mathbf{f}_g (\mathbf{q}, \dot{\mathbf{q}})$. To do this, we use the pull-back operator to transform both $\mathbf{f}_r(\mathbf{x}_r, \dot{\mathbf{x}}_r)$ and $\mathbf{M_r}(\mathbf{x}_r, \dot{\mathbf{x}}_r)$ from task-space $\mathcal{R}$ into $\mathcal{Q}$ as follows:
\begin{equation*}
   ^{\mathcal{Q}}\mathbf{f}_r = 
    \text{pull}_{\mathcal{Q}}\left(
    \mathbf{f}_r(\mathbf{x}_r, \dot{\mathbf{x}}_r)\right) = (\mathbf{J}_r^{\top} \mathbf{M}_r \mathbf{J}_r)^{\dagger} \mathbf{J}_r^{\top} \mathbf{M}_r \mathbf{f}_r
\end{equation*}
\begin{equation*}
   ^{\mathcal{Q}}\mathbf{M}_r = 
    \text{pull}_{\mathcal{Q}}\left(
    \mathbf{M}_r(\mathbf{x}_r, \dot{\mathbf{x}}_r)\right) = \mathbf{J}_r^{\top} \mathbf{M}_r \mathbf{J}_r
\end{equation*}

\textbf{Combining Policies.} Given the goal-attracting policy $\mathbf{f}_g$ and the dynamic obstacle-avoiding policy transformed in joint-space $^{\mathcal{Q}}\mathbf{f}_r$, we can combine these two policies to yield the full DCP-RMP as follows:
\begin{equation*}
    \ddot{\mathbf{q}}_{mg}(t) = 
    (^{\mathcal{Q}}\mathbf{M}_r + \mathbf{M}_g)^{\dagger} \
    (^{\mathcal{Q}}\mathbf{M}_r ^{\mathcal{Q}}\mathbf{f}_r + \mathbf{M}_g \mathbf{f}_g)
\end{equation*}
Finally, we perform Euler integration to yield the modified joint goal $\mathbf{q}_{md}$. We note that in practice, we perform multiple Euler integration steps per control-loop iteration to result in a more reactive behavior. In essence, $\mathbf{q}_{mg}$ prioritizes reactive dynamic obstacle avoidance if obstacles move close to the robot; otherwise, $\mathbf{q}_{mg}$ prioritizes reaching towards the desired joint position goal $\mathbf{q}_g$. 
\begin{equation*}
    \mathbf{q}_{mg} (t+1) =  \text{EulerIntegrate}\left(\mathbf{q}_{mg}(t), \dot{\mathbf{q}}_{mg}(t), \ddot{\mathbf{q}}_{mg}(t)\right)
\end{equation*}

\section{Architecture Ablation}
\label{arch-ablation}
We compare IMPACT to the LSTM-GMM architecture used in the state-of-the-art neural motion policy NeuralMP~\cite{nmp}, holding all other factors— from dataset to pretraining and finetuning—constant to isolate the effect of architecture. As shown in Table~\ref{tab:ablate-archi}, the LSTM-GMM (from NeuralMP) improves performance by 25.5\% over its original baseline when trained with our data and finetuning scheme. However, replacing it with IMPACT adds a \textbf{further 9\% gain}. Furthermore, prior results from NeuralMP show that NeuralMP outperforms M$\pi$Nets and other baselines when trained on the same dataset. These findings confirm that our improvements stem not only from data and training, but also from our architectural design.

\begin{table}[h]
    \centering
    \begin{tabular}{@{}lcccccc@{}}
        \toprule  & SE & SAO & FDO & GB & DGB \\
        \midrule
        NeuralMP~\cite{nmp} & 75.56 & 76.00 & 57.50 & 37.73  & 57.00\\
        IMPACT & \textbf{84.60} & \textbf{86.00} & \textbf{75.50} & \textbf{66.67} & \textbf{65.25} \\
        \bottomrule
    \end{tabular}
    \vspace{0.4em}
    \caption{Architecture: IMPACT vs. NeuralMP.}\label{tab:ablate-archi}
\end{table}

\section{Network Size and Inference Speed Comparison}
Our network runs at 300 Hz on an RTX 4090 GPU and AMD 7950X CPU. We report comparisons of the network size and Cold-Start (CS) time—used in the M$\pi$Nets paper to measure reactivity—as the average time to respond to a new planning problem, in Tab. \ref{tab:size_and_speed}.

\begin{table}[h]
    \centering
    \footnotesize 
    \setlength{\tabcolsep}{3pt}
    \begin{tabular}{l@{\hskip 4pt} c@{\hskip 4pt} c@{\hskip 4pt} c@{\hskip 4pt} c@{\hskip 4pt} c@{\hskip 4pt} c}
        \toprule
        & \textbf{DRP} & NeuralMP (with TTO) & M$\pi$Nets & M$\pi$Former \\
        \midrule
        Network Parameter Size & \textbf{4.67M} & 20M & 19M & 26M \\
        Cold-Start Time (ms) & \textbf{3.48} & 2970  & 6.8 & 30  \\
        \bottomrule
    \end{tabular}
    \vspace{0.4em}
    \caption{Network size and cold-start time comparisons.}
    \label{tab:size_and_speed}
\end{table}

\section{Evaluation Details}
Here we provide a detailed breakdown of \textbf{goal-reaching (R)} and \textbf{collision (C)} rates on \textbf{DRPBench} for all methods.
\begin{table}[h]
    \centering
    \setlength{\tabcolsep}{10pt}
    \begin{tabular}{@{}lccccc@{\hspace{1pt}}c@{}}
    \toprule
    & \multicolumn{5}{c}{\textbf{DRPBench (R/C)}}\\
    \cmidrule(lr){2-6}
    & SE & SAO & FDO & GB & DGB & \\
    \midrule
    \multicolumn{7}{@{}l}{\textit{With privileged information}:} \\
    AIT* & 60.81/20.31 & - & - & - & -\\
    cuRobo & 91.71/10.66 & 94.00/36/33 & 93.00/60.00 & 0.00/23.46 & 91.33/96.75\\
    \midrule
    cuRobo-Vox & 79.25/45.35 & 96.00/38.67 & 96.50/59.00 & 0.00/21.25 & 91.50/97.25\\
    RMP & 38.69/49.88 & 54.33/35.00 & 55.50/21.00 & \textbf{88.19/22.77} & 54.00/24.00\\
    \midrule
    M$\pi$Nets & 3.25/46.56 & 0.67/47.00 & 1.00/68.00 & 2.21/71.15 & 1.75/95.25\\
    M$\pi$Former & 0.41/56.72 & 0.00/44.00 & 0.00/53.00 & 0.00/23.28 & 0.25/58.00\\
    NeuralMP & 81.97/49.06 & 86.33/44.33 & 97.00/80.50 & 97.94/100.00 & 92.75/99.50\\
    \midrule
    IMPACT & \textbf{91.34/10.16} & \textbf{95.67/12.33} & 97.00/67.50 & 95.00/31.64 & 95.00/99.25\\
    \textit{DRP (Ours)} & \textbf{91.34/10.16} & \textbf{95.67/12.33} & \textbf{96.00/23.00} & 95.00/31.64 & \textbf{95.50/32.75}\\
    \bottomrule
    \end{tabular}
    \vspace{0.4em}
    \caption{Reaching success rate on DRPBench.}\label{tab:sim_results}
\end{table}

\section{Detailed Model Parameters}

We list key hyperparameters for training IMPACT in Table~\ref{tab:hyperparam}.

\begin{table}[h]
\centering
\begin{tabular}{@{}p{5cm}p{4cm}@{}}
\toprule
\textbf{Hyperparameter} & \textbf{Value} \\
\midrule
\multicolumn{2}{c}{\textbf{Input Parameters}} \\
\midrule
\# Scene PCD ($N_s$) & 2048 \\
\# Robot PCD ($N_r$) & 256 \\
\midrule
\multicolumn{2}{c}{\textbf{IMPACT Pre-training Parameters}} \\
\midrule
Optimizer & AdamW \\
Base Learning Rate & 0.001 \\
Weight Decay & 0.01 \\
Optimizer Momentum & $\beta_1, \beta_2 = 0.9, 0.95$ \\
Batch Size & 512 \\
Learning Rate Schedule & Cosine Decay \\
Total Steps & 1,000,000 \\
Warmup Steps & 100 \\
GPU & RTX4090 (24 gb) \\
Wall-Clock Time & 3 days\\
\midrule
\multicolumn{2}{c}{\textbf{IMPACT Fine-tuning Parameters}} \\
\midrule
Optimizer & AdamW \\
Base Learning Rate & 0.001 \\
Weight Decay & 0.01 \\
Optimizer Momentum & $\beta_1, \beta_2 = 0.9, 0.95$ \\
Batch Size & 512 \\
\# Updates for Base Model & 5 \\
GPU & RTX4090 (24 gb) \\
Wall-Clock Time & 10 hours\\
\midrule
\multicolumn{2}{c}{\textbf{IMPACT Architecture}} \\
\midrule
\multicolumn{2}{c}{\textbf{PCD Set Abstraction Layer}} \\
Radius & 0.1 \\
\# Samples & 64 \\
MLP Hidden Layers & $[64,64,64]$ \\
\midrule
\multicolumn{2}{c}{\textbf{Current/Target Joint Configuration Encoder}} \\
MLP Hidden Layers & $[128, 256]$ \\
\midrule
\multicolumn{2}{c}{\textbf{Transformer Parameters}} \\
\# Scene PCD Tokens ($K_s$) & 128 \\
\# Robot PCD Tokens ($K_r$) & 16 \\
\# Encoder Layers & 6 \\
\# Decoder Layers & 6 \\
\# MHSA Heads & 8 \\
Hidden Dim & 128 \\
Feed-Forward Dim & 1024 \\
Positional Encoding & sin cos \\
Action Chunk & 10 \\
\bottomrule
\end{tabular}
\vspace{2mm}
\caption{\small Policy Architecture and Training Hyperparameters}\label{tab:hyperparam}
\vspace{-4mm}
\end{table}

\end{document}